# Polarized Skylight Navigation System (PSNS)/ GNSS/SINS Integrated Navigation

Huaju Liang, and Hongyang Bai

*Abstract*—Bioinspired polarized skylight navigation system (PSNS) is a new navigation method, which imitates insects using polarized skylight for navigation. However, the latest research shows that the current PSNS is difficult to obtain the three-dimensional (3D) attitude in real time only relying on polarized skylight. This causes not only that three-dimensional (3D) cannot be estimated only by PSNS in real time, but also when polarized light is integrated with other navigation methods, 3D attitude cannot be directly used for integrated navigation. Therefore, an integrated navigation algorithm based on bi-direction solar vector is proposed, and a DSP+FPGA (digital signal processor + field programmable gate array) dual-core architecture PSNS/GNSS/SINS (global navigation satellite system/strapdown inertial navigation system) integrated navigation system was constructed.

## I. Introduction

Many insects can extract heading angle from the polarized skylight, such as locusts, desert ants and so on [1, 2]. There are special regions in the compound eyes of these insects, which are called dorsal rim area (DRA) and can perceive the skylight polarization patterns. Then, polarized skylight signals from compound eyes are integrated in the central complex for navigation. And the skylight polarization patterns are relatively stable and contain rich information. Therefore, polarized skylight navigation has great research value and application prospects. Besides, the polarized light navigation is not interfered by human factors, has good concealment and real-time performance. So polarized skylight navigation will be an ideal way to navigate in the future.

Since the discovery that some insects can navigate by sensing the polarized skylight, there are more and more researches on bioinspired polarization skylight navigation. At present, several heading estimation methods (one-dimensional attitude algorithms) based on PSNS have been proposed and applied to the navigation of aircraft, ships, vehicles, robots and so on [3-8]. Moreover, based on least square method, two-dimensional attitude algorithms have been proposed by obtaining the solar vector from skylight polarization patterns [9-11]. However, the latest research shows that the current PSNS is difficult to estimate three Euler angles in real time only relying on polarized skylight [12]. So, based only on polarized light navigation, it cannot meet the requirements of current three-dimensional navigation. In addition, when polarized skylight navigation is integrated with other navigation methods, three-dimensional attitude cannot be directly used for integration.

So, at present, a variety of integrated navigation systems with PSNS aided have been proposed, which promote the application of PSNS. Most these methods only use the orientation obtained by PSNS to construct integrated navigation system [11, 13], so the information of polarized skylight is not fully utilized. The latest researches began to consider the value of solar vector in integrated navigation [14, 15]. However, all these methods do not pay attention to the fact that the solar vector obtained by PSNS is a bi-direction vector. In other words, it is difficult to distinguish whether the vector captured by PSNS is solar vector or anti-solar vector.

Aiming at the problem in the current polarized skylight integrated navigation systems, a PSNS/GNSS (global navigation satellite system) /SINS integrated navigation algorithm is designed based on bi-direction solar vector. This algorithm integrated PSNS, GNSS and SINS based on Kalman filtering, and the attitude angles are not directly integrated in this algorithm, but the bi-direction solar vector is selected as the observed quantity to modify the integrated attitude. In addition, a PSNS/GNSS/SINS integrated navigation system based on DSP + FPGA (digital signal processor + field programmable gate array) dual-core architecture is designed to improve the performance of the navigation system, where DSP is the signal processing core and FPGA is the data transmission core.

In short, this paper proposes a PSNS/GNSS/SINS integrated navigation algorithm based on bi-direction solar vector, and constructs a DSP + FPGA dual core architecture integrated navigation system.

## II. Integrated Navigation Algorithm

A PSNS/GNSS/SINS integrated navigation algorithm is designed in this section. Current PSNS is difficult to obtain the three-dimensional attitude directly, but can capture bi-direction solar vector. So, this paper proposes the integrated navigation algorithm using bi-direction solar vector captured by PSNS. The flow diagram of the integrated navigation algorithm is

This work was supported in part by the National Natural Science Foundation of China under Grant 61603189. (Corresponding author: Hongyang Bai.)

H. Liang and H. Bai are with the School of Energy and Power Engineering, Nanjing University of Science and Technology (NJUST), Nanjing 210094, China (e-mail: lianghuaju@sina.cn ; hongyang@njust.edu.cn)

shown in Fig. 1. It should be noted that the sun model can obtain solar vector in term of position and time [16-18]. But the solar vector captured by PSNS is a bi-direction solar vector, that is, it is difficult to distinguish whether it is solar vector or anti-solar vector.

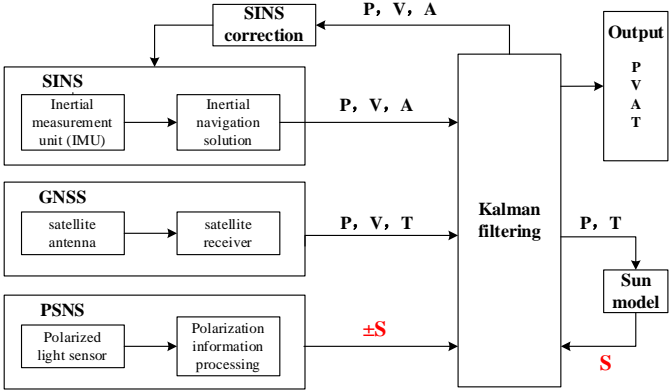

Fig. 1. PSNS/GNSS/SINS integrated navigation algorithm, where P represents position, V represents velocity, A represents attitude, T represents time, S represents solar vector, ±S represents bi-direction solar vector.

*A. Bi-direction solar vector captured by PSNS*
*1) Measurement of polarized electric field vectors*

Polarized light imager can get four direction light intensity information $I_0$, $I_{45}$, $I_{90}$, and $I_{135}$ in one shot, where subscripts of 0, 45, 90, and 135 are the directions of polarizers. Stokes vector $(S_0, S_1, S_2, S_3)$ is calculated to describe polarization state.

$$\begin{cases} S_0 = \frac{1}{2}[I_0 + I_{45} + I_{90} + I_{135}] \\ S_1 = I_0 - I_{90} \\ S_2 = I_{45} - I_{135} \\ S_3 = 0 \end{cases} \quad (1)$$

where $S_3 = 0$ represents that circularly polarized light is very rare in the sky. Then, degree of polarization (DOP) and angle of polarization (AOP) are given by

$$DOP = \frac{\sqrt{S_1^2 + S_2^2 + S_3^2}}{S_0} \quad (2)$$

$$AOP = \begin{cases} \frac{1}{2}\arctan\left(\frac{S_2}{S_1}\right) & (S_1 > 0) \\ \frac{1}{2}\arctan\left(\frac{S_2}{S_1}\right) + 90° & (S_1 < 0) \wedge (S_2 > 0) \\ \frac{1}{2}\arctan\left(\frac{S_2}{S_1}\right) - 90° & (S_1 < 0) \wedge (S_2 < 0) \end{cases} \quad (3)$$

Polarized light imager can easily obtain DOP and AOP by Stokes vector. However, polarized electric field vectors (E-vectors) cannot be obtained directly.

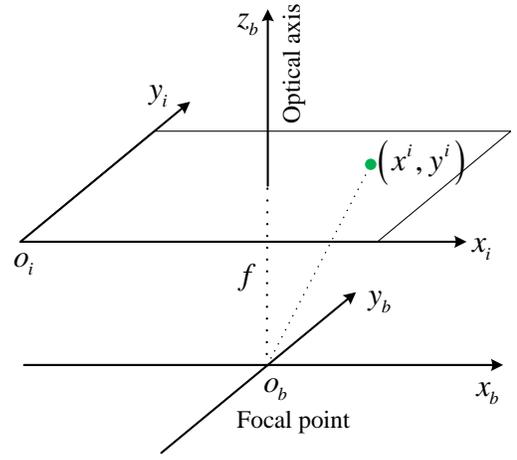

Fig. 2. Body coordinate system $o_b x_b y_b z_b$ and image plane coordinate system $o_i x_i y_i$, where $f$ is focal length.

So, image plane coordinate system is constructed with the x-axis ($x_i$) and y-axis ($y_i$) along the column and row directions of image plane, respectively. And the unit for this coordinate system is the pixel. $(x^i, y^i)$ represents a pixel in image plane coordinate system, Then, the observed direction of pixel $(x^i, y^i)$ in body coordinate system is

$$V_P^b = \left(D_x(x^i - \frac{\eta_x+1}{2}), D_y(y^i - \frac{\eta_y+1}{2}), f\right)^T \quad (4)$$

where $D_x$ and $D_y$ are the pixel size in $x_i$ and $y_i$ directions, respectively. $\eta_x$ and $\eta_y$ the number of pixels in $x_i$ and $y_i$ directions, respectively. $f$ represents the focal length of polarized light imager.

AOP captured by polarized light imager is the angle between $y_b$ and the projection of polarized E-vector in image plane. So, the projection of polarized E-vector in image plane is

$$\pm E_P^i = \pm(\sin AOP, \cos AOP, 0)^T \quad (5)$$

So, suppose the polarized E-vector in body coordinate system is

$$\pm E_P^b = \pm(\sin AOP, \cos AOP, E_{PZ}^b)^T \quad (6)$$

Since the observed polarized E-vector is located on the celestial sphere, the polarized E-vector is always perpendicular to the observed direction. So

$$\pm E_P^b \cdot V_P^b = 0 \quad (7)$$

Then, we can get

$$E_{PZ}^b = \frac{D_x(x^i - \frac{\eta_x+1}{2})\sin AOP + D_y(y^i - \frac{\eta_y+1}{2})\cos AOP}{f} \quad (8)$$

Above all, polarized E-vectors are obtained based on PSNS and bi-direction solar vector can be captured according to equation (6).

*2) Measurement of Bi-direction solar vector*

According to Rayleigh sky model, all the polarized E-vectors in the sky are perpendicular solar vector. Then we can get

$$\pm E_P^b \cdot S^b = 0 \tag{9}$$

where $\pm E_P^b$ and $S^b$ represent the polarized E-vectors and solar vector in body coordinate system, respectively. $\pm$ represents the polarized E-vectors are bi-direction vectors, the subscript $P$ represents the observation point $P$, the superscript $b$ represents body coordinate system. As shown in Fig. 2, the body coordinate system $o_b x_b y_b z_b$ has its origin $o_b$ at the center of polarized light imager with the z-axis ($z_b$) pointing along the optical axis of the polarized light imager, the x-axis ($x_b$) and y-axis ($y_b$) along the column and row directions of image plane, respectively.

In addition, because of the symmetry of Rayleigh sky model, anti-solar vector also satisfies

$$\pm E_P^b \cdot (-S^b) = 0 \tag{10}$$

where $-S^b$ represents anti-solar vector in body coordinate system. So, we can get

$$\pm E_P^b \cdot \pm S^b = 0 \tag{11}$$

where $\pm S^b$ represents bi-direction solar vector (solar vector and anti-solar vector) in body coordinate system. Then, $\pm S^b$ can be obtained by least square method. The error function is defined as

$$error = \sum_{i=1}^n \omega_i \left(\pm E_{Pi}^b \cdot \pm S^b\right)^2 - \lambda \left(\left\|\pm S^b\right\|^2 - 1\right) \tag{12}$$

where, $\omega_i$ is the weight coefficient, which is proportional to the degree of polarization (DOP). $-\lambda \left(\left\|\pm S^b\right\|^2 - 1\right)$ is the constraint that $\pm S^b$ is a unit vector, and $\lambda \geq 0$. $n$ represents the total number of observation points. Then we get

$$\begin{aligned} error &= \sum_{i=1}^n \omega_i \left(\pm S^b\right)^T \cdot \pm E_{Pi}^b \cdot \left(\pm E_{Pi}^b\right)^T \cdot \pm S^b - \lambda \left(\pm S^b\right)^T \cdot \pm S^b + \lambda \\ &= \sum_{i=1}^n \omega_i \left(\pm S^b\right)^T \cdot \left[E_{Pi}^b \cdot \left(E_{Pi}^b\right)^T\right] \cdot \pm S^b - \lambda \left(\pm S^b\right)^T \cdot \pm S^b + \lambda \end{aligned} \tag{13}$$

According to the least square method, when (13) takes the minimum,

$$\lambda \cdot \pm S^b = \sum_{i=1}^n \omega_i \left[E_{Pi}^b \cdot \left(E_{Pi}^b\right)^T\right] \cdot \pm S^b = K \cdot \pm S^b \tag{14}$$

According to equation (14), $\lambda$ is the eigenvalue of matrix K, and $\pm S^b$ is the eigenvector corresponding to the minimum eigenvalue. Above all, the eigenvector corresponding to the minimum eigenvalue is bi-direction solar vector $\pm S^b$.

### B. State equation

The PSNS/GNSS/SINS integrated navigation system takes the ENU (east, north, and up) coordinate system as the navigation system, and the estimated quantity X is

$$X = [\varphi_E, \varphi_N, \varphi_U, \delta v_E, \delta v_N, \delta v_U, \delta L, \delta \lambda, \delta h,$$
$$\varepsilon_x, \varepsilon_y, \varepsilon_z, \nabla_x, \nabla_y, \nabla_z]^T \tag{15}$$

where the subscripts $E$, $N$, $U$ represent east, north, and up respectively, the subscripts $x$, $y$ and $z$ represent $x_b$, $y_b$ and $z_b$ axes of body coordinate system. $\varphi_E$, $\varphi_N$, and $\varphi_U$ represent attitude errors; $\delta v_E$, $\delta v_N$, and $\delta v_U$ represent velocity errors; $\delta L$, $\delta \lambda$, and $\delta h$ represent latitude error, longitude error, and height error, respectively; $\varepsilon_x$, $\varepsilon_y$, $\varepsilon_z$, represent the drifts of the gyro. $\nabla_x$, $\nabla_y$, and $\nabla_z$ represent the bias of the accelerometer.

According to the given the estimated quantity $X$, the state equation of this integrated navigation system is given by

$$\dot{X}(t) = F(t)X(t) + G(t)W(t) \tag{16}$$

where $F(t)$ represents state transfer matrix, $G(t)$ represents noise transfer matrix, $W(t)$ represents noise vector.

The state transfer matrix $F(t)$ of this navigation system is

$$F(t) = \begin{bmatrix} F_N & F_S \\ 0_{6 \times 9} & F_M \end{bmatrix}_{15 \times 15} \tag{17}$$

where $F_N$ is the $9 \times 9$ dimensional basic navigation parameter system, and its nonzero elements can be find in [19]. $F_S$ and $F_M$ is given by

$$F_S = \begin{bmatrix} C_b^n & 0_{3 \times 3} \\ 0_{3 \times 3} & C_b^n \\ 0_{3 \times 3} & 0_{3 \times 3} \end{bmatrix}_{9 \times 6} \tag{18}$$

$$F_M = 0_{6 \times 6} \tag{19}$$

where $C_b^n$ is the transformation matrix from body coordinate system to ENU coordinate system calculated by the mathematical platform.

The noise transfer matrix $G(t)$ of this navigation system is

$$G(t) = \begin{bmatrix} C_b^n & 0_{3 \times 3} \\ 0_{3 \times 3} & C_b^n \\ 0_{9 \times 3} & 0_{9 \times 3} \end{bmatrix}_{15 \times 6} \tag{20}$$

### C. Measurement equation

The measurement equation of the designed PSNS/GNSS/SINS integrated navigation system are based on position measurement equation, velocity measurement equation, bi-direction solar vector measurement equation.

*1) Position measurement equation*

The output position of SINS is

$$\begin{cases} L_s = L + \delta L \\ \lambda_s = \lambda + \delta \lambda \\ h_s = h + \delta h \end{cases} \tag{21}$$

where $L_s$, $\lambda_s$ and $h_s$ represent latitude, longitude, and height measured by SINS. $L$, $\lambda$, $h$ represent the true position of the carrier. $\delta L$, $\delta \lambda$ and $\delta h$ represent latitude, longitude, and height differences between true position and the position measured by SINS.

The output position of GNSS is

$$\begin{cases} L_g = L - N_E / R_{MH} \\ \lambda_g = \lambda - N_N / (R_{NH} \cos L) \\ h_g = h - N_U \end{cases} \tag{22}$$

where $L_g$, $\lambda_g$ and $h_g$ represent latitude, longitude, and height, measured by GNSS. $N_E$, $N_N$ and $N_U$ are the east, north, and up directions position errors of the GNSS. $R_{MH}$ is the meridian radius of curvature along the north-south direction. $R_{NH}$ the radius of curvature of the prime vertical along east-west direction.

The position measurement vector $Z_P$ is given based on the difference of the output position of SINS and GNSS.

$$Z_P = \begin{bmatrix} R_{MH}\delta L + N_E \\ R_{NH}\cos L \delta\lambda + N_N \\ \delta h + N_U \end{bmatrix} = H_P X + V_P \quad (23)$$

where

$$H_P = \begin{bmatrix} 0_{3\times 6} & diag(R_{MH} & R_{NH}\cos L & 1) & 0_{3\times 6} \end{bmatrix} \quad (24)$$

$$V_P = \begin{bmatrix} N_E & N_N & N_U \end{bmatrix} \quad (25)$$

*2) Velocity measurement equation*

The output velocity of SINS is

$$\begin{cases} V_{sE} = V_E + \delta V_E \\ V_{sN} = V_N + \delta V_N \\ V_{sU} = V_U + \delta V_U \end{cases} \quad (26)$$

where $V_{sE}$, $V_{sN}$ and $V_{sU}$ represent the velocity measured by SINS. $V_E$, $V_N$, $V_U$ represent the true velocity of the carrier. $\delta V_E$, $\delta V_N$, $\delta V_U$ represent the differences between true velocity and the velocity measured by SINS.

The output velocity of GNSS is

$$\begin{cases} V_{gE} = V_E - M_E \\ V_{gN} = V_N - M_N \\ V_{gU} = V_U - M_U \end{cases} \quad (27)$$

where $V_{gE}$, $V_{gN}$ and $V_{gU}$ represent the velocity measured by GNSS. $M_E$, $M_N$, and $M_U$ are east, north, and up directions velocity errors of the GNSS.

The velocity measurement equation is given based on the difference of the output velocity of SINS and GNSS.

$$Z_V = \begin{bmatrix} \delta V_E + M_N \\ \delta V_E + M_E \\ \delta V_E + M_U \end{bmatrix} = H_V X + V_V \quad (28)$$

where

$$H_V = \begin{bmatrix} 0_{3\times 3} & diag(1 & 1 & 1) & 0_{3\times 9} \end{bmatrix} \quad (29)$$

$$V_V = \begin{bmatrix} M_E & M_N & M_U \end{bmatrix} \quad (30)$$

*3) Bi-direction solar vector measurement equation*

The relationship between the solar vector obtained by sun position model and the solar vector measured by PSNS is

$$S^n = \hat{C}_b^n (S^b + \xi_p) \quad (31)$$

where $S^n$ represents the solar vector in ENU coordinate system captured by sun position model, where $\hat{C}_b^n$ is the true transformation matrix from body coordinate system to ENU coordinate system, $S^b$ represents the solar vector in body coordinate system captured by PSNS, $\xi_p$ is the noise error of the polarized light sensors, $(S^b + \xi_p)$ represents the true solar vector in body coordinate system.

Special attention should be paid to the fact that the sun vector obtained by PSNS is a bi-directional vector, so (31) is rewritten.

$$S^n \cong \hat{C}_b^n (\pm S^b + \xi_p) \quad (32)$$

where $\pm S^b$ represents the bi-direction solar vector in body coordinate system captured by PSNS, $\cong$ represents one direction of $\pm S^b$ makes the equation true.

Due to the existence of mathematical platform misalignment angle $\phi^n$, the relationship between the measured transformation matrix is $C_b^{n1}$ and true transformation matrix $\hat{C}_b^n$ is

$$\hat{C}_b^n = C_{b1}^n C_b^n \quad (33)$$

where $C_{b1}^n$ represents attitude error matrix. When the error angle $\phi^n = [\phi_E \ \phi_N \ \phi_U]^T$ is small, the high-order term can be ignored and the simplified attitude error matrix can be obtained.

$$C_{b1}^n = I + (\phi\times) = \begin{bmatrix} 1 & -\phi_U & \phi_N \\ \phi_U & 1 & -\phi_E \\ -\phi_E & \phi_E & 1 \end{bmatrix} \quad (34)$$

Then,

$$\hat{C}_b^n = C_{b1}^n C_b^n = (I + (\phi\times))C_b^n \quad (35)$$

Substituting (32) into (29), we can get

$$S^n \cong (I + (\phi\times))C_b^n (\pm S^b + \xi_p) \quad (36)$$

Let $\delta C = (\phi\times)$, then we can get it

$$S^n - C_b^n (\pm S^b) \cong (I + \delta C)C_b^n \xi_p + \delta C \cdot C_b^n (\pm S^b) \quad (37)$$

where $\delta C$ is the cross product antisymmetric matrix of the attitude error angle $\phi$, so we can get

$$S^n - C_b^n (\pm S^b) \cong -((C_b^n (\pm S^b))\times) \cdot \phi + (I + \delta C)C_b^n \xi_p \quad (38)$$

For the left side of the equation (38), it represents the measurement error of the solar vector. Because of the assumption that the measurement error is small, so for $S^n - C_b^n (+S^b)$ and $S^n - C_b^n (-S^b)$, the one with smaller vector modulus is the result.

In bi-direction solar vector measurement equation, the attitude matrix measured by SINS is used. And let $A = C_b^n S^b$, $\tilde{A} = A\times$ is the cross product anti-symmetric matrix of A, which satisfies:

$$\tilde{A} = A\times = \begin{bmatrix} 0 & -A_3 & A_2 \\ A_3 & 0 & -A_1 \\ -A_2 & A_1 & 0 \end{bmatrix} \quad (39)$$

Then, according to (38), the solar vector measurement vector $Z_S$ is given by

$$Z_S = \begin{bmatrix} S_E + A_E \\ S_N + A_N \\ S_U + A_U \end{bmatrix} = H_S X + V_S \quad (40)$$

where

$$H_S = \begin{bmatrix} 0 & A_3 & -A_2 & 0_{1\times 12} \\ -A_3 & 0 & A_1 & 0_{1\times 12} \\ A_2 & -A_1 & 0 & 0_{1\times 12} \end{bmatrix} \quad (41)$$

$$V_S = \begin{bmatrix} Q_E & Q_N & Q_U \end{bmatrix} \quad (42)$$

where $Q_E$, $Q_N$, and $Q_U$ are solar vector errors of the PSNS along east, north, and up directions.

*4) System measurement equation*

Above all, combine velocity, position and solar vector measurement equations, and the measurement equation of the designed PSNS/GNSS/SINS integrated navigation is

$$Z = \begin{bmatrix} Z_V \\ Z_P \\ Z_S \end{bmatrix} = \begin{bmatrix} H_V \\ H_P \\ H_S \end{bmatrix} X + \begin{bmatrix} V_V \\ V_P \\ V_S \end{bmatrix} \quad (43)$$

### D. Discrete Kalman Filter

As mentioned above, the state equation (16) and measurement equation (43) of PSNS/GNSS/SINS integrated algorithm are in a continuous time form. In this section, equations (16) and (43) are rewrite in a discrete-time form to make it easier to program.

$$\begin{cases} X_k = \Phi_{k,k-1} X_{k-1} + \Gamma_{k-1} W_{k-1} \\ Z_k = H_k X_k + V_k \end{cases} \quad (44)$$

where $\Phi_{k,k-1}$ represents transfer matrix; $\Gamma_{k-1}$ represents system noise drive matrix; $X_{k-1}$ represents system state on last moment; $W_k$ represents system noise sequence. $H_k$ represents system measurement matrix; $V_k$ represents measurement noise sequence;

The update steps discrete Kalman filter is shown below.

state prediction:

$$\hat{X}_{k|k-1} = \Phi_{k,k-1} \hat{X}_{k-1} \quad (45)$$

where $\hat{X}_{k-1}$ is state estimation at time of $k-1$; $\hat{X}_{k|k-1}$ is the estimate of $X$ at time $k$ given observations up to and including at time $k-1$;

State estimation:

$$\hat{X}_k = \hat{X}_{k|k-1} + K_k (Z_k - H_k \hat{X}_{k|k-1}) \quad (46)$$

where $\hat{X}_k$ is state estimation at time of $k-1$; $K_k$ is the gain matrices at time of $k-1$.

The optimal Kalman gain matrices is

$$K_k = P_{k|k-1} H_k^T (H_k P_{k|k-1} H_k^T + R_k)^{-1} \quad (47)$$

where $R_k$ is the covariance of the observation noise. Predicted estimate covariance matrix $P_{k|k-1}$ is given by

$$P_{k|k-1} = \Phi_{k,k-1} P_{k-1} \Phi_{k,k-1}^T + \Gamma_{k-1} Q_{k-1} \Gamma_{k-1}^T \quad (48)$$

where $Q_{k-1}$ is the covariance of the process noise. And a posteriori estimate covariance matrix is

$$P_k = (I - K_k) P_{k|k-1} \quad (49)$$

where $I$ is a unit matrix.

## III. INTEGRATED NAVIGATION SYSTEM

As shown in Fig. 3, the designed PSNS/GNSS/SINS integrated navigation system is mainly composed of polarization imager, inertial measurement unit (IMU), GNSS receiver and dual-core missile-borne computer. The specific hardware composition and parameters of this system are as follows.

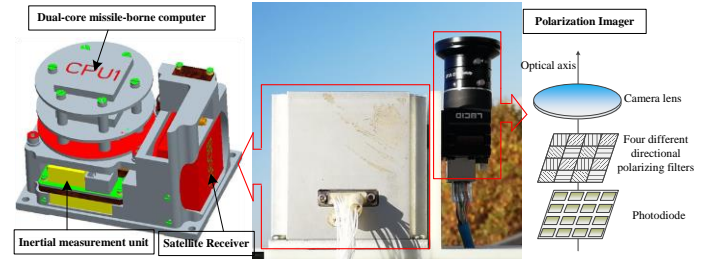

Fig. 3. Mainly composed of PSNS/GNSS/SINS integrated navigation system.

### A. PSNS

PSNS is mainly composed of Sony PHX050S-P polarization imager. The imaging unit of this polarization imager is Sony IMX250MZR COMS (Complementary Metal Oxide Semiconductor) whose resolution is 2448 * 2048, pixel size is 3.45um, format is 2/3". Especially, the IMX250MZR COMS has four different polarization directional (0°, 90°, 45°, and 135°) polarizing filters on every four pixels, as shown in Fig. 3. So, it can capture the light intensity information of four polarization directions in real time to calculate degree and angle of polarization in real time.

### B. GNSS

GNSS is mainly composed of satellite antenna and satellite receiver, whose horizontal position error is less than 10m, altitude positioning error is less than 15m, horizontal velocity measurement accuracy is 0.1m/s, altitude velocity error is less than 0.2m/s, second pulse accuracy is up to 50ns, loss of lock recapture time is less than 1s, data output frequency can be set to 1Hz.

### C. SINS

SINS is mainly composed of three-axis Sensonor STIM210 MEMS (Micro-Electro-Mechanical System) gyroscope, Colibrys MS9030 and V8015 MEMS accelerometers. Its data output frequency is 200Hz; the measured zero bias repeatability of the gyroscope is less than 40° / h, the bias stability is less than 20 ° / h, the nonlinearity is less than 0.02% FS, the output symmetry is less than 0.05%, and the measurement is ±400° / S; the bias repeatability of the x-axis accelerometer MS9030 is less than 7mg, the bias stability is less than 3mg, the scale factor stability is less than 900ppm, the scale error is less than 200ppm, and the measurement range is up to The results show that the repeatability of bias is less than 7mg, the bias stability is less than 4 mg, the scale factor stability is less than 1000 ppm, the scale error is less than 200ppm, and the range is ± 25 G.

### D. Dual-core missile-borne computer

The dual-core missile-borne computer is mainly composed of TMS320C6747 DSP Central Processing Unit (CPU) and Sparten-3e FPGA. TMS320C6747 DSP CPU is the signal processing core, and responsible for PSNS, GNSS and SINS signal processing, integrated navigation calculation and other tasks. Sparten-3e FPGA is the data transmission core to realize the fast data transmission, which is responsible for data acquisition of each sensor and monitoring data transmission. It has the characteristics of high speed, high precision, low power consumption, small size, rich peripheral modules and interfaces. Its operation speed is as high as 300MHz, its processing speed is higher than 1200MFLOPS,

The dual-core architecture of this system is shown in Fig. 4. The output of three-axis gyroscope is digital signal, and the interface is RS422. The voltage signal of accelerometers is converted into digital signal output by A/D conversion circuit. After filtering and full temperature compensation for the collected data of gyro and accelerometer, the data is sent to FPGA through RS422 serial port. The polarization imager captures the skylight polarization patterns and obtains the solar vector, which is transmitted to FPGA through RS232. Satellite antenna receives satellite information to satellite receiver to obtain satellite navigation information. Then, it is sent to FPGA through RS422 serial port.

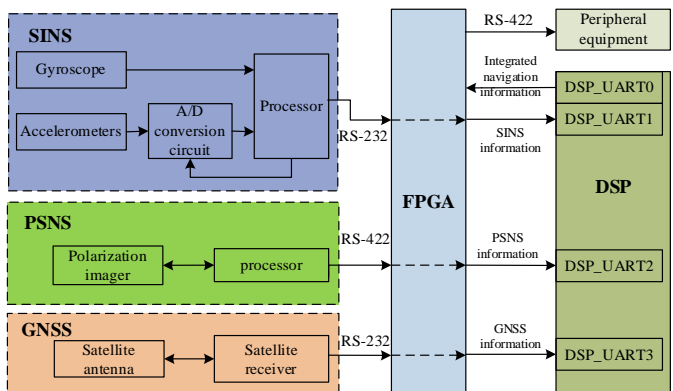

Fig. 4. Dual-core architecture of PSNS/GNSS/SINS integrated navigation system.

TMS320C6747 DSP has four serial ports, which are DSP_UART0, DSP_UART1, DSP_UART2 and DSP_UART3, and it extends the isolated multiplex standard full-duplex serial port through the FPGA. SINS is connected to DSP_UART1 through RS422 serial port of FPGA. SINS is connected to DSP_UART1 through RS232 serial port of FPGA. PSNS is connected to DSP_UART2 through RS422 serial port of FPGA. GNSS is connected to DSP_UART3 through RS232 serial port of FPGA. DSP_UART0 is mapped to RS422 serial port to communicate with peripheral devices through FPGA.

## IV. CONCLUSION

The major contribution of this paper is that an PSNS/GNSS/SINS integrated navigation algorithm based on bi-direction solar vector is proposed, and a DSP+FPGA dual-core architecture PSNS/GNSS/SINS integrated navigation system was constructed. We are preparing related experiments to verify the effectiveness of the algorithm and the system.

In this paper, Kalman filtering algorithm is applied to integrated navigation algorithm. However, other filtering algorithms can also be applied to this system, and we will carry out related researches in the future.